\documentclass[11pt,a4paper]{article}
\usepackage{times,latexsym}
\usepackage{url}
\usepackage[T1]{fontenc}
\usepackage{times}
\usepackage{latexsym}
\usepackage{amsmath}
\usepackage{multirow}
\usepackage{url}
\usepackage{tipa}
\usepackage{arydshln}
\usepackage{tikz}
\usepackage{tikz-qtree} 
\usepackage{bussproofs}
\usepackage{stmaryrd}
\usepackage{multicol} 
\usepackage{graphics}  
\usepackage{adjustbox}
\usepackage{linguex} 
\usepackage{soul}
\usepackage{pbox}
\usepackage{amsmath}
\usepackage{amssymb}
\usepackage{pifont}
\usepackage{outlines}
\usepackage{stackrel}
\usepackage{caption}
\usepackage{xcolor}
\usepackage{csquotes}
\usepackage{multirow}
\usepackage{array}
\newcolumntype{P}[1]{>{\arraybackslash}p{#1}}
\usepackage{booktabs}
\usepackage{url}
\usepackage{fancyhdr}
\newcommand{\cmark}{\ding{51}}%
\usepackage[font=small]{caption}
\makeatletter
\newcommand{\@BIBLABEL}{\@emptybiblabel}
\newcommand{\@emptybiblabel}[1]{}
\makeatother
\usepackage[hidelinks]{hyperref}

\newcommand{\newterm}[1]{\textit{#1}} % textsc takes up a lot of space. Replaced all textsc with this, so we can change it back later if we want to.

\usepackage[acceptedWithA]{tacl2018v2}

\usepackage{xspace,mfirstuc,tabulary}

\newif\iftaclinstructions
\taclinstructionsfalse % AUTHORS: do NOT set this to true
\iftaclinstructions
\renewcommand{\confidential}{}
\renewcommand{\anonsubtext}{(No author info supplied here, for consistency with
TACL-submission anonymization requirements)}
\newcommand{\instr}
\fi

\iftaclpubformat % this "if" is set by the choice of options

\else

\fi

\title{Neural Network Acceptability Judgments}

\author{
Alex Warstadt\\
New York University\\
\texttt{\small warstadt@nyu.edu}
\And
Amanpreet Singh \\
New York University\\
Facebook AI Research\thanks{\hspace{0.5em}Current affiliation. This work was completed when the author was at New York University.}\\
\texttt{\small amanpreet@nyu.edu}
\And
Samuel R.~Bowman\\
New York University\\
\texttt{\small bowman@nyu.edu}\\
} 

\date{}

\begin{document}
\maketitle
\begin{abstract}
This paper investigates the ability of artificial neural networks to judge the grammatical acceptability of a sentence, with the goal of testing their linguistic competence.  
We introduce the Corpus of Linguistic Acceptability (CoLA), a set of 10,657 English sentences labeled as grammatical or ungrammatical from published linguistics literature.
As baselines, we train several recurrent neural network models on acceptability classification, and find that our models outperform unsupervised models by Lau et al.~(2016) on CoLA. 
Error-analysis on specific grammatical phenomena reveals that both Lau et al.'s models and ours learn systematic generalizations like subject-verb-object order.
However, all models we test perform far below human level on a wide range of grammatical constructions. 
\end{abstract}

\section{Introduction}

Artificial neural networks (ANNs) achieve a high degree of competence on many applied natural language understanding (NLU) tasks, but this does not entail that they have knowledge of grammar. A key property of a human's linguistic competence is the ability to identify in one's native language, without formal training in grammar, a contrast in acceptability\footnote{Following terminological conventions in linguistics, a sentence's \newterm{grammaticality} is determined by a grammatical formalism, while its \newterm{acceptability} is determined by introspective judgments of native speakers \cite{schutze1996empirical}.} between pairs of sentences like those in \ref{island}. \newterm{Acceptability judgments} like these are the primary behavioral measure that generative linguists use to observe humans' grammatical knowledge \cite{chomsky1957syntactic,schutze1996empirical}.

\setlength{\Exlabelsep}{0em}
\setlength{\SubExleftmargin}{1.5em}
\ex.\label{island}\a.What did Betsy paint a picture of?
\b.*What was a picture of painted by Betsy?

We train neural networks to perform acceptability judgments---following work by \newcite{lawrence2000grammatical}, \newcite{lau2016cognitive}, and others---in order to evaluate their acquisition of the kinds of grammatical concepts linguists identify as central to human linguistic competence. This contributes to a growing effort to test ANNs' ability to make fine-grained grammatical distinctions \cite{linzen2016assessing,adi2017fine,conneau2018cram,ettinger2018assessing,marvin2018targeted}. This research program seeks to provide new informative ways to evaluate ANN models popular with engineers. Furthermore, it has the potential to address foundational questions in theoretical linguistics by investigating how well unbiased learners can acquire grammatical knowledge.

\begin{table*}[t]
\centering \small

\begin{tabular}{l l l l}
\toprule
\multirow{3}{*}{Included}&
Morphological Violation&(a)&*Maryann should leaving.\\
&Syntactic Violation&(b)&*What did Bill buy potatoes and \underline{~~}?\\
&Semantic Violation&(c)&*Kim persuaded it to rain.\\
\midrule
\multirow{4}{*}{Excluded}&
Pragmatical Anomalies&(d)&
*Bill fell off the ladder in an hour.\\
&Unavailable Meanings&(e)&
*He$_i$ loves John$_i$. (\emph{intended}: John loves himself.)\\
&Prescriptive Rules&(f)&
\phantom{*}Prepositions are good to end sentences with.\\
&Nonce Words&(g)&
*This train is arrivable.\\
\bottomrule
\end{tabular}

\caption{Our informal classification of unnacceptable sentences, shown with their presence or absence in CoLA.}\label{include exclude}
\end{table*}

In this paper we make four concrete contributions: (i) We introduce the Corpus of Linguistic Acceptability (CoLA), a collection of sentences from the linguistics literature with expert acceptability labels which, at over 10k examples, is by far the largest of its kind. (ii) We train several semi-supervised neural sequence models to do acceptability classification on CoLA and compare their performance with unsupervised models from \newcite{lau2016cognitive}. Our best model outperforms unsupervised baselines, but falls short of human performance on CoLA by a wide margin. (iii) We analyze the impact of supervised training on acceptability classifiers by varying the domain and quantity of training data. (iv) We assess our models' performance on acceptability classification of specific linguistic phenomena. These experiments illustrate how acceptability classification and CoLA can give detailed insights into what grammatical knowledge typical neural network models can acquire. We find that our models do not show evidence of learning non-local dependencies related to agreement and questions, but do appear to acquire knowledge about basic subject-verb-object word order and verbal argument structure.

\paragraph{Resources}

% CoLA will be available upon publication through a dedicated corpus site, which will also include source code and an interactive demo for our best model. We will also make available a scoring site for evaluating acceptability classifiers on our private test data.

CoLA can be downloaded from the corpus website.\footnote{\href{https://nyu-mll.github.io/CoLA/}{\url{https://nyu-mll.github.io/CoLA/}}}\label{links_start} The code for training our baselines is available as well.\footnote{\href{https://github.com/nyu-mll/CoLA-baselines}{\url{https://github.com/nyu-mll/CoLA-baselines}}} There are also two competition sites for evaluating acceptability classifiers on CoLA's in-domain\footnote{\href{https://www.kaggle.com/c/cola-in-domain-open-evaluation}{\url{https://www.kaggle.com/c/cola-in-domain-open-evaluation}}} and out-of-domain\footnote{\href{https://www.kaggle.com/c/cola-out-of-domain-open-evaluation}{\url{https://www.kaggle.com/c/cola-out-of-domain-open-evaluation}}} test sets (unlabeled). Finally, CoLA is included in the GLUE benchmark\footnote{\href{https://gluebenchmark.com/tasks}{\url{https://gluebenchmark.com/tasks}}}\label{links_end} \citep{wang2018glue}, which also hosts CoLA training data, (unlabeled) test data, and a leaderboard.

\section{Acceptability Judgments}\label{acceptability judgments}

\subsection{In Linguistics}

Our investigation of acceptability classification builds on decades of established scientific knowledge in generative linguistics, where acceptability judgments are studied extensively. In his foundational work on generative syntax, \newcite{chomsky1957syntactic} defines an empirically adequate grammar of a language $L$ as one which generates all and only those strings of $L$ which native speakers of $L$ judge to be acceptable.
Evaluating grammatical theories against native speaker judgments has been the dominant paradigm for research in generative syntax over the last sixty years \cite{schutze1996empirical}. Linguists generally provide evidence in the text of their papers in the form of constructed example sentences annotated with Boolean acceptability judgments from themselves or native speakers.

\subsection{The Acceptability Classification Task}

While acceptability classification has been explored previously in computational linguistics, there is no standard approach to this task. Following common practice in generative linguistics our study focuses on the Boolean acceptability classification task. This approach is also taken in earlier computational work on this task \cite{lawrence2000grammatical,wagner2009grammaticality,linzen2016assessing}. By contrast, other computational work aims to model gradient acceptability judgments \cite{heilman2014grammaticality,lau2016cognitive}. Though \citeauthor{lau2016cognitive}\ argue that acceptability judgments are gradient in nature, we consider Boolean judgments in published examples sufficient for our purposes, since linguists generally design these examples to be unambiguously acceptable or unacceptable.

Datasets for acceptability classification require a source of unacceptable sentences, which are not generally found in naturalistic speech or writing by native speakers.
The sentences in CoLA consist entirely of examples from the linguistics literature. \newcite{lawrence2000grammatical} and \newcite{lau2016cognitive} build datasets similar in this respect. However, at over 10k sentences, CoLA is by far the largest dataset of this kind, and represents the widest range of sources. Prior work in this area also obtains unacceptable sentences by programmatically generating fake sentences that are unlikely to be acceptable.
\newcite{wagner2009grammaticality} distort real sentences by, for example, deleting words, inserting words, or altering
verbal inflection. \newcite{lau2016cognitive} use round-trip
machine-translation from English into various languages and back.
% \citet{heilman2014grammaticality} source sentences from essays by non-native English speakers. 
We also generate fake sentences to pre-train our baselines before further training on CoLA.

We see several advantages in using linguistics example sentences. First, they are labeled for acceptability by the authors, thereby simplifying the annotation process. Second, because linguists present examples to motivate arguments, these sentences isolate a particular grammatical construction while minimizing superfluous content. Hence, unacceptable sentences in CoLA tend to be maximally similar to acceptable sentences and are unacceptable for a single identifiable reason.

We note that \newcite{gibson2010quantitative} express concern about standard practices around acceptability judgments and call for theoretical linguists to quantitatively measure the reliability of the judgments they report, sparking an ongoing dialog about the validity and reproducibility of these judgments \cite{sprouse2012adger,sprouse2017setting,sprouse2013li,mahowald2016snap}. 
We take no position on this general question, but perform a small human evaluation to gauge the reproducibility of the judgments in CoLA (Section \ref{human performance}).

\subsection{The Role of Minimal Pairs}\label{sec:minimal pairs}

Acceptability judgments can alternatively be framed as a forced choice between \emph{minimal pairs}, i.e.~pairs of minimally different sentences contrasting in acceptability as in \ref{island}, where the classifier or subject selects the sentence with greater (predicted) acceptability. This kind of judgment has been taken as a standard for replicability of reported judgments in syntax articles \citep{sprouse2012adger,sprouse2013li,linzen2018reliability}. It is also increasingly used in computational linguistics \citep{linzen2016assessing,marvin2018targeted,futrell2018rnns,wilcox2018rnn,wilcox2019structural}. This task is often employed to evaluate language models because the outputted probabilities for a pair of minimally different sentences are directly comparable, while the output for a single sentence cannot be taken as a measure of acceptability without some kind of normalization \citep{lau2016cognitive}.

We leave a comparison of this methodology with our own for future work. We settle on the single-sentence judgment task because it is directly comparable with methodology in generative linguistics. While some work in theoretical linguists presents acceptability judgments as a ranking of two or more sentences \citep[][pp.~77-81]{schutze1996empirical}, Boolean judgments are still the norm, and the dominant current theories still make Boolean \emph{predictions} about whether a sentence is or is not grammatical \cite[][pp.~12-16]{chomsky1995minimalist}. Accordingly, CoLA, but not datasets based solely on preferences between minimal pairs, may be used to evaluate models' ability to make judgments that align with both native speaker judgments and the predictions of generative theories.

\subsection{Defining (Un)acceptability}\label{defining}

Not all linguistics examples are suitable for acceptability classification. While all acceptable sentences can be included, we exclude four types of unacceptable sentences from the task (examples in Table \ref{include exclude}):

\paragraph{Pragmatic Anomalies} Examples like (d) are interpretable, but in odd scenarios distinguishable from plausible scenarios only with access to real-world knowledge unrelated to grammar.

\paragraph{Unavailable Meanings} Examples like (e) are often used to illustrate that a sentence cannot express a particular meaning. This example can only express that someone other than John loves John. We exclude these examples because there is no simple way to force an acceptability classifier to consider only the interpretation in question.

\paragraph{Prescriptive Rules} Examples like (f) violate rules which are generally explicitly taught rather than being learned naturally, and are therefore not considered a part of native speaker grammatical knowledge in linguistic theory.

\paragraph{Nonce Words} Examples like (g) illustrate impossible affixation or lexical gaps. Since these words will not appear in the vocabularies of typical word-level NLP models, they will be impossible for these models to judge.

The acceptability judgment task as we define it still requires identifying challenging grammatical contrasts. A successful model needs to recognize (a) morphological anomalies such as mismatches in verbal inflection, (b) syntactic anomalies such as wh-movement out of extraction islands, and (c) semantic anomalies such as violations of animacy requirements of verbal arguments.

\section{CoLA}\label{cola}

This paper introduces the Corpus of Linguistic Acceptability (CoLA),
%\footnote{CoLA can be downloaded here:\\ Removed for anonymity.}
a set of example sentences from the linguistics literature labeled for acceptability. CoLA is available online, alongside source code for our baseline models, and a leaderboard showing model performance on test data using privately-held labels (see footnotes 2-6 for links).

\paragraph{Sources}

We compile CoLA with the aim of representing a wide variety of phenomena of interest in theoretical linguistics. We draw examples from linguistics publications spanning a wide time period, a broad set of topics, and a range of target audiences. Table \ref{sources} enumerates our sources.
By way of illustration, consider the three largest sources in the corpus: Kim \& Sells \shortcite{kim2008syntax} is a recent undergraduate syntax textbook, Levin \shortcite{levin1993verb} is a comprehensive reference detailing the lexical properties of thousands of verbs, and Ross \shortcite{ross1967constraints} is an influential dissertation focusing on wh-movement and extraction islands in English syntax.

\begin{table}[t!]
\centering
\scriptsize
\begin{tabular}{>{\raggedright\hangindent=.7em\hangafter=1}p{.18\textwidth}rr>{\raggedright\hangindent=.7em\hangafter=1}p{.12\textwidth}}
\toprule
\textbf{Source}&\textbf{N}&\textbf{\%}
&\textbf{Topic}\tabularnewline
\midrule
\newcite{adger2003core}&948&71.9&Syntax Textbook\tabularnewline
\newcite{baltin1982landing}&96&66.7&Movement\tabularnewline
\newcite{baltin2001handbook} &880&66.7&Handbook\tabularnewline
\newcite{bresnan1973comparative}&259&69.1&Comparatives\tabularnewline
\newcite{carnie2013syntax}&870&80.3&Syntax Textbook\tabularnewline
\newcite{culicover1999comparative}&233&59.2&Comparatives\tabularnewline
\newcite{dayal1998inherently}&179&75.4&Modality\tabularnewline
\newcite{gazdar1981unbounded}&110&65.5&Coordination\tabularnewline
\newcite{goldberg2004resultative}&106&77.4&Resultative\tabularnewline
\newcite{kadmon1993any}&93&81.7&Negative Polarity\tabularnewline
\newcite{kim2008syntax}&1965&71.2&Syntax Textbook\tabularnewline
\newcite{levin1993verb}&1459&69.0&Verb alternations\tabularnewline
\newcite{miller2002syntax}&426&84.5&Syntax Textbook\tabularnewline
\newcite{hovav2008dative}&151&69.5&Dative alternation\tabularnewline
\newcite{ross1967constraints}&1029&61.8&Islands\tabularnewline
\newcite{sag1985coordination}&153&68.6&Coordination\tabularnewline
\newcite{sportiche2013introduction}&651&70.4&Syntax Textbook\tabularnewline
\textbf{In-Domain}&\textbf{9515}&\textbf{71.3}&\tabularnewline
\midrule
\newcite{chung1995sluicing}&148&66.9&Sluicing\tabularnewline
\newcite{collins2005smuggling}&66&68.2&Passive\tabularnewline
\newcite{jackendoff1971gapping}&94&67.0&Gapping\tabularnewline
\newcite{sag1997relativeclause}&112&57.1&Relative clauses\tabularnewline
\newcite{sag2003syntactic}&460&70.9&Syntax Textbook\tabularnewline
\newcite{williams1980predication}&169&76.3&Predication\tabularnewline
\textbf{Out-of-Domain}&\textbf{1049}&\textbf{69.2}&\tabularnewline
\midrule
\textbf{Total}&\textbf{10657}&\textbf{70.5}&\tabularnewline
\bottomrule 
\end{tabular}
\caption{The contents of CoLA by source. \textit{N} is the number of sentences in a source. \textit{\%} is the percent of sentences labeled acceptable. Sources listed above \textit{In-Domain} are included in the training, development, and test sets, while those above  \textit{Out-of-Domain} appear only in the development and test sets.}\label{sources}
\end{table}

\begin{table*}[t] 
\centering
\footnotesize
% SB: I randomly removed a few for space.
% TODO: How did you sort these? Say something in the caption, as the order (by author, but not alphabetical) is a bit odd.
\begin{tabular}{l >{\raggedright\hangindent=1em\hangafter=1}p{.58\textwidth} l}
\toprule \textbf{Label}&\textbf{Sentence}&\textbf{Source}\\\midrule
*&The more books I ask to whom he will give, the more he reads. & \newcite{culicover1999comparative}\\
\cmark&I said that my father, he was tight as a hoot-owl. & \newcite{ross1967constraints}\\
\cmark&The jeweller inscribed the ring with the name. & \newcite{levin1993verb}\\
*&many evidence was provided.&\newcite{kim2008syntax}\\
\cmark&They can sing.&\newcite{kim2008syntax}\\
\cmark&The men would have been all working.&\newcite{baltin1982landing}\\
*&Who do you think that will question Seamus first?&\newcite{carnie2013syntax}\\
*&Usually, any lion is majestic.&\newcite{dayal1998inherently}\\
\cmark&The gardener planted roses in the garden.&\newcite{miller2002syntax}\\
\cmark&I wrote Blair a letter, but I tore it up before I sent it.&\newcite{hovav2008dative}\\
\bottomrule
\end{tabular}
\caption{CoLA random sample, drawn from the in-domain training set (\cmark = acceptable, *=unacceptable).}
\end{table*}

\paragraph{Preparing the Data}

The corpus includes all usable examples from each source. We manually remove unacceptable examples falling into any of the excluded categories described in Section \ref{defining}.
The labels in the corpus are the original authors' acceptability judgments whenever possible. When examples appear with non-Boolean judgments (this occurs in less than 3\% of cases), we either exclude them (for labels `?' or `\#'), or label them unacceptable (`??' and `*?'). We also expand examples with optional or alternate phrases into multiple data points, e.g.~\emph{Betsy buttered (*at) the toast} becomes \emph{Betsy buttered the toast} and \emph{*Betsy buttered at the toast}.

In some cases, we change the content of examples slightly. To avoid irrelevant complications from out-of-vocabulary words, we restrict CoLA to the 100k most frequent words in the British National Corpus, and edit sentences as needed to remove words outside that set. For example, \emph{That new handle unscrews easily} is replaced with \emph{That new handle detaches easily} to avoid the out-of-vocabulary word \emph{unscrews}. We make these alterations manually to preserve the author's stated intent, in this case selecting another verb that undergoes the middle voice alternation.

Finally, we define acceptability classification as a sentence classification task. To ensure that all examples in CoLA are sentences, we augment fragmentary examples, replacing, for example, \emph{*The Bill's book} with \emph{*The Bill's book has a red cover}.

\paragraph{Splitting the Data}

In addition to the train/development/test split used to control overfitting in standard benchmark datasets, CoLA is further divided into an in-domain set and an out-of-domain set, as specified in Table \ref{sources}. The out-of-domain set is constructed to be about 10\% the size of CoLA and to include sources of varying sizes, degrees of domain specificity, and time period.\footnote{In Section \ref{sec:cola design} we consider several alternate splits of CoLA.} The in-domain set is split three ways into training (8551 examples), development (527), and test sets (530), all drawn from the same 17 sources. The out-of-domain set is split into development (516) and a test sets (533), drawn from another 6 sources. We split CoLA in this way in order to monitor two types of overfitting during training: overfitting to the specific sentences in the training set (in-domain), and overfitting to the specific sources and phenomena represented in the training set (out-of-domain).

\paragraph{Phenomena in CoLA}

\begin{figure}
    \centering
    \includegraphics[width=0.5\textwidth]{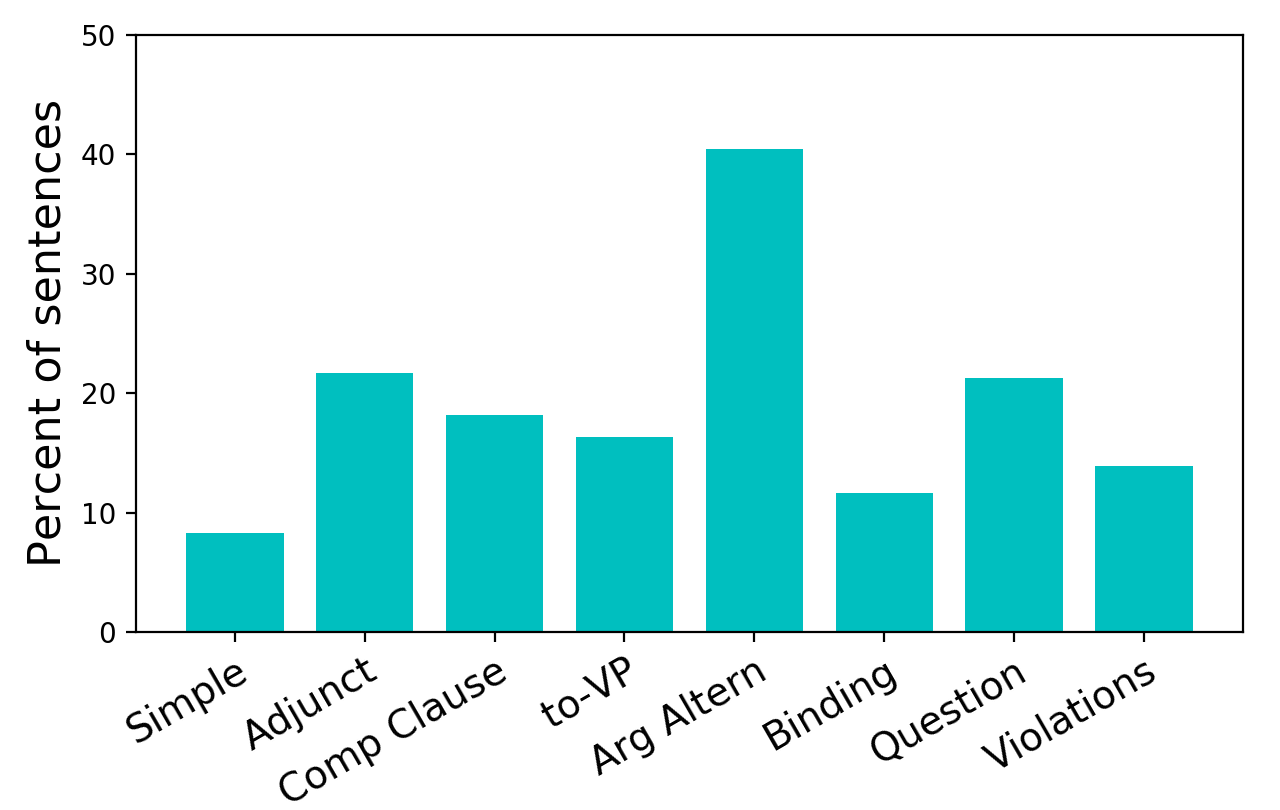}
    \caption{Frequencies of phenomenon types in the CoLA development set.}
    \label{fig:freqs}
\end{figure}

CoLA has wide coverage of syntactic and semantic phenomena. To quantify the distribution of phenomena represented, we annotate the entire CoLA development set for the presence of constructions falling into 15 broad classes, of which 8 are discussed here, for brevity.\footnote{The annotated data also includes 63 fine-grained features. The annotated data is available for download on the CoLA website alongside detailed annotation guidelines.} Briefly, \emph{simple} labels sentences with no marked syntactic structures; \emph{adjunct} labels sentences that contain adjuncts of nouns and verb phrases; \emph{comp clause} labels sentences with embedded or complement clauses; \emph{to-VP} labels sentences with non-finite embedded verb phrase; \emph{arg altern} labels sentences with non-canonical argument structures such as passives; \emph{binding} labels sentences with pronouns and binding phenomena; \emph{question} labels sentences with interrogative clauses and relative clauses; and \emph{violations} labels sentences with morphological or semantic violations, or an extra/missing word. The average sentence is labeled with 3.22 features.

Figure \ref{fig:freqs} shows the frequency of these 8 features in the development set. Argument alternations are the best represented phenomenon and appear in over 40\% of sentences in this sample. This is due both to the high frequency of these constructions as well as the inclusion of several sources directly addressing this topic \citep{levin1993verb,collins2005smuggling,hovav2008dative}. Most other constructions appear in about 10-20\% of sentences, indicating that CoLA is fairly balanced according to this annotation scheme. There are likely biases in CoLA that other annotation schemes could detect. However, it is open to debate what a balanced dataset for acceptability judgments should look like. There is no agreed upon set of key phenomena in linguistics and any attempt to create one is likely to be controversial and overly simplistic. Furthermore, if such a set of phenomena did exist, the builders of a balanced dataset must decide whether it should be balanced equally across phenomena, or weighted by either the frequency in broad coverage corpora of English or the number of distinguishing syntactic contrasts associated with each phenomenon. We assume that CoLA skews towards the latter, as a major goal of linguistics articles is to document key unique facts about some phenomenon without excessive repetition.

\paragraph{Human Performance}\label{human performance}

We measure human performance on a subset of CoLA to set an approximate upper bound for machine performance on acceptability classification and to estimate the reproducibility of the judgments in CoLA. We have five linguistics PhD students, all native English speakers, perform a forced-choice single-sentence acceptability judgment task on 200 sentences from CoLA, divided evenly between the in-domain and out-of-domain development sets. These human judgments are available alongside on the corpus site. 

Results appear in Table \ref{results table}. Average annotator agreement with CoLA is 86.1\%, and average Matthews Correlation Coefficient (MCC)\footnote{MCC \cite{matthews1975correlation} is an evaluation metric for unbalanced binary classifiers. It is a special case of Pearson's $r$ for Boolean variables, i.e.\ it measures correlation of two Boolean distributions, giving a value between -1 and 1. On average, any two unrelated distributions will have an MCC of 0, regardless of class imbalance. By contrast, accuracy and F1 favor classifiers with a majority-class bias.} is 0.697.
Selecting the majority decision from our annotators gives us a rough upper bound on human performance. These judgments agreed with CoLA's ratings on 87\% of sentences with an MCC of 0.713. In other words, 13\% of the labels in CoLA contradict the observed majority judgment. 

We identify several reasons for disagreements between our annotators and CoLA. Errors in character recognition in the source PDFs may produce artifacts which alter the acceptability of the sentence or omit the original judgment. Based on these 200 sampled sentences, we estimate such errors occur in 1-2\% of CoLA sentences. Ascribing two percentage points of disagreement to such errors, the remaining eleven points can be ascribed to a lack of context or genuine variation between the dialect spoken by the original author and that spoken by the annotator.\footnote{We observe greater disagreement between human annotators and published judgments than \newcite{sprouse2013li} do. As a reviewer points out, this may be due to the fact that \citeauthor{sprouse2013li} measure agreement with minimal pairs of sentences using a forced choice task, which is more constrained and arguably easier than single sentence judgments.} We also measure our \textit{individual} annotators' agreement with the aggregate rating, yielding an average pairwise agreement of 93\%, and an average MCC of 0.852. 

\section{Experiments}\label{modeling}

We train several semi-supervised neural network models to do acceptability classification on CoLA. At 10k sentences, CoLA is likely too small to train a low-bias learner like a recurrent neural network  without additional prior knowledge. In similar low-resource
settings, transfer learning with sentence embeddings
has proven to be effective \cite{kiros2015skip,conneau2017supervised}. Our best model uses a transfer learning approach in which a large sentence encoder is trained on an unsupervised real/fake discrimination objective, and a lightweight multilayer perceptron classifier is trained on top to do acceptability classification over CoLA. It also uses contexualized word embeddings inspired by ELMo \cite{peters2018elmo}. 

We compare our models to a continuous bag of words (CBOW) baseline, the unsupervised models proposed by \cite{lau2016cognitive}, and human performance. To make these comparisons more meaningful, we avoid giving our models distinct advantages over human learners by limiting the training data in two ways: (i) Aside from acceptability labels, our training has no grammatical annotation. (ii) Our large sentence encoders are limited to 100-200 million tokens of training data, which is within a factor of ten of the number of tokens human learners are exposed to during language acquisition \cite{hart1992parenting}.\footnote{\newcite{hart1992parenting} find that children in affluent families
are exposed to about 45 million tokens by age 4.} We avoid training models on significantly more data because such models have a distinct advantage over the human learners we aim to match.

\subsection{Preliminaries}

\paragraph{Language Model}

We use an LSTM language model (LSTM LM) at various stages in our experiments: (i) Several of our models use word embeddings or hidden states from the LM as input. (ii) The LM generates fake data for the real/fake task. (iii) The LM is an integral part of our implementation of the method proposed by \cite{lau2016cognitive}. We train the LM on the 100 million-token British National Corpus (BNC). It learns word embeddings from scratch for the 100k most frequent words in the BNC (with out of vocabulary words replaced by $<$\texttt{unk}$>$). We lowercase and tokenize the BNC data using NLTK \cite{bird2004nltk}. The LM achieves a word-level perplexity of 56.1 on the BNC.

\paragraph{Word Representations}

We experiment with three styles of word representations: (i) We train a set of conventional fixed word embeddings as part of the training of the LM described above, which we refer to as \emph{BNC embeddings}. (ii) We train \newterm{ELMo-style} contextualized word embeddings, which, following ELMo \cite{peters2018elmo}, represent $w_i$ as a linear combination of the hidden states $h_i^j$ for each layer $j$ in an LSTM LM, though we depart from the original paper by using only a forward LM. (iii) We also use the pretrained 300-dimensional (6B) \newterm{GloVe embeddings} from \citet{pennington2014glove}.\footnote{Results with models that use these GloVe embeddings are less immediately comparable with human performance results, since GloVe is trained on several orders of magnitude more text than humans see during language acquisition.}

\paragraph{Real/Fake Auxiliary Task}

We train sentence encoders on a \newterm{real/fake task} in which the objective is to distinguish real sentences from the BNC and ``fake'' English sentences automatically generated by two strategies: (i) We sample strings, e.g.~\ref{lm}, from the LSTM LM. (ii) We manipulate sentences of the BNC, e.g.~\ref{shuff}, by randomly permuting a subset of the words, keeping the other words \emph{in situ}. Training data includes the entire BNC and an equal amount of fake data. We lowercase and tokenize all real/fake data and replace out of vocabulary words  as in LM training.

\ex.
\a.\label{lm} either excessive tenure does not threaten a value to death.
\b.\label{shuff} what happened in to the empire early the traditional roman portrait?

We choose this task because arbitrary numbers of labeled fake sentences can be generated without using any explicit knowledge of grammar in the process, and we expect that many of the same features are relevant to both the real/fake task and the downstream acceptability task. 

\begin{figure}[t]\centering
\includegraphics[width=0.83\columnwidth]{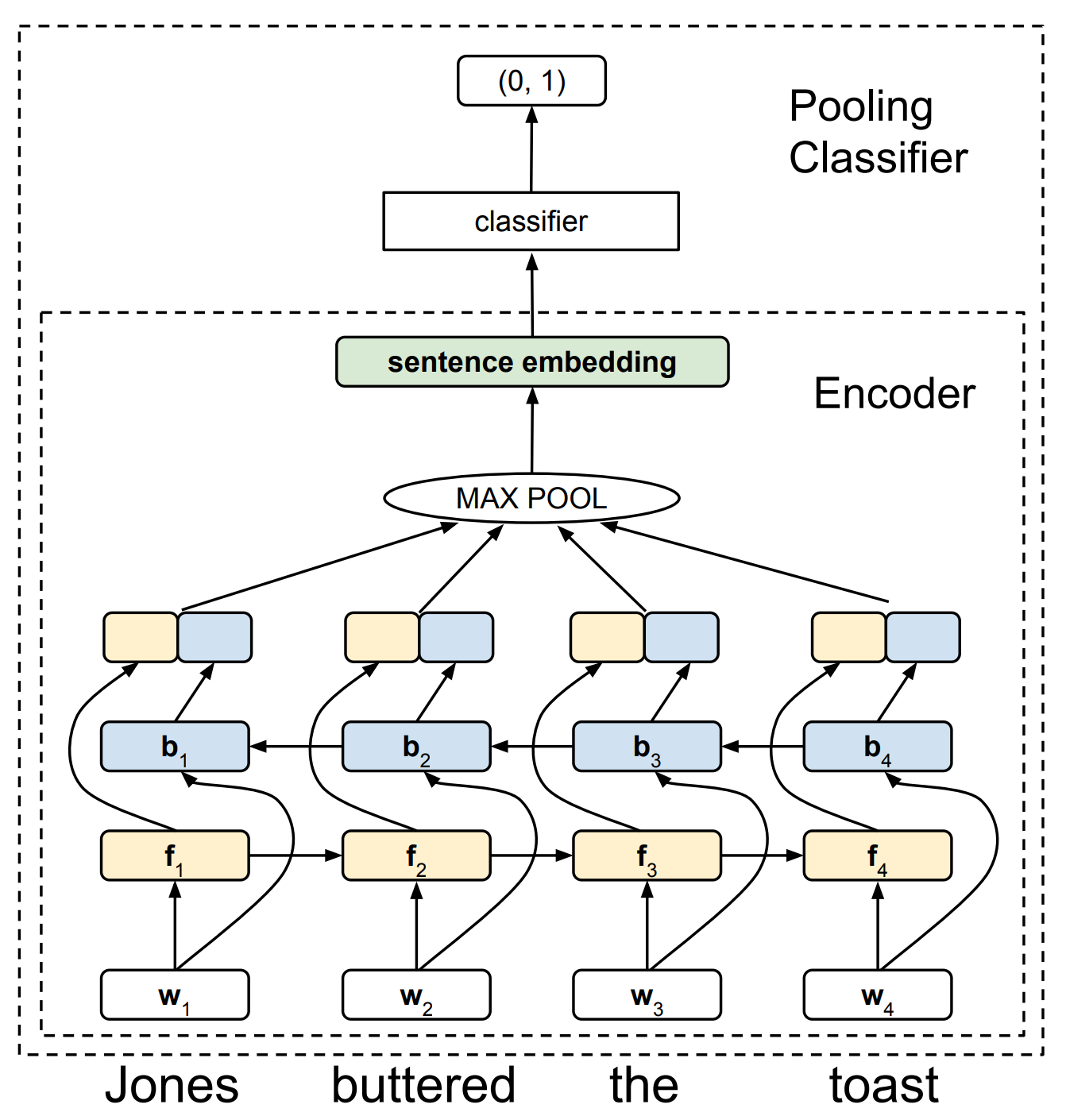}
\caption{Architecture for the pooling classifier models. $w_i$ = word embeddings, $f_i$ = forward LSTM hidden state, $b_i$ = backward LSTM hidden state.}\label{real/fake}
\end{figure}

\subsection{Baselines}

\begin{table*}[t]
\centering \small
\begin{tabular}{llllrrrr}
\toprule

\multirow{2}{*}{\textbf{Model}}&
\multirow{2}{*}{\textbf{Embeddings}}&
\multirow{2}{*}{\textbf{\pbox{0.5in}{Encoder\\Training}}}&
\multirow{2}{*}{\textbf{\pbox{0.5in}{Classifier\\ Training}}}&
\multicolumn{2}{c}{\textbf{In-domain}}&
\multicolumn{2}{c}{\textbf{Out-of-domain}}\\
&&&&\textbf{Acc.}
&\textbf{MCC}
&\textbf{Acc.}
&\textbf{MCC}
\\\midrule
CBOW
&BNC
&--
&CoLA
&0.502
&0.063
&0.482
&0.096 \\\midrule
LSTM LM WLPM
&BNC
&--
&CoLA Thresh.
&0.652
&0.253
&0.711
&0.238
\\
%
% 4-gram LM SLOR
% &--
% &--
% &CoLA Thresh.
% &0.642
% &0.223
% &0.645
% &0.042\\
% %
% 3-gram LM SLOR
% &--
% &--
% &CoLA Thresh.
% &0.646
% &0.212
% &0.681
% &0.141\\
% %
% 2-gram LM SLOR
% &--
% &--
% &CoLA Thresh.
% &0.590
% &0.162
% &0.707
% &0.180\\
%
4-gram LM WLPM
&--
&--
&CoLA Thresh.
&0.474
&0.000
&0.645
&0.042\\
3-gram LM WLPM
&--
&--
&CoLA Thresh.
&0.428
&0.142
&0.681
&0.141\\
2-gram LM WLPM
&--
&--
&CoLA Thresh.
&0.452
&0.094
&0.707
&0.180\\\midrule
Pooling Classifier
&BNC
&Real/Fake
&Real/Fake&
0.728&
0.196&
0.707&
0.180
\\
Pooling Classifier
&GloVe
&Real/Fake
&Real/Fake&
0.766&
0.302&
0.660&
0.063
\\
Pooling Classifier
&ELMo-Style
&Real/Fake
&Real/Fake&
0.758&
0.265&
0.702&
0.177
\\\midrule
Pooling Classifier
&ELMo-Style
&CoLA
&CoLA&
0.726&
0.278&
0.651&
0.155\\
Pooling Classifier
&BNC
&Real/Fake
&CoLA
&0.723
&0.261
&0.679
&0.186
\\
Pooling Classifier
&GloVe
&Real/Fake
&CoLA
&0.706
&0.300
&0.608
&0.135\\
Pooling Classifier
&ELMo-Style
&Real/Fake
&CoLA&
\textbf{0.772}&
\textbf{0.341}&
\textbf{0.732}&
\textbf{0.281}
\\\midrule
Human Average&
--&--&--&
\em 0.850&
\em 0.644&
\em 0.872&
\em 0.738\\
Human Aggregate&
--&--&--&
\em 0.870&
\em 0.695&
\em 0.910&
\em 0.815\\
\bottomrule
\end{tabular}
\caption{Results for acceptability classification on the CoLA test set. The first group is the CBOW baseline. The second group is the \textit{LSTM} and $n$\textit{-gram} LMs with Lau et al.'s metrics. The third group is pooling classifiers trained end-to-end on the real/fake objective. The fourth group is pooling classifiers with training on CoLA, mostly with encoders transferred from real/fake classifiers. The fifth group is the small human evaluations (Section \ref{human performance}). \emph{CoLA-Thresh.} is threshold tuning on CoLA, and \emph{WLPM} is Lau et al.'s Word LogProb Min-1 metric.}\label{results table}

\end{table*}

\paragraph{Pooling Classifier}

Our real/fake classifiers and acceptability classifiers use an architecture we refer to as a \newterm{pooling classifier} which is based on \newcite{conneau2017supervised}. As illustrated in Figure \ref{real/fake}, the pooling classifier consists of two parts: (i) a sentence encoder which reduces variable-length sequences of tokens into fixed-length \newterm{sentence embeddings}, and (ii) a lightweight classifier which outputs a classification based on the sentence embedding. In the sentence encoder, a deep bidirectional LSTM reads a sequence of word embeddings; then the forward and backward hidden states for each time step are concatenated, and max-pooling over the sequence gives a sentence embedding. In the classifier, the sentence embedding is passed through a sigmoid output layer (optionally preceded by a single hidden layer) giving a scalar representing the probability of a positive classification (either the sentence is real or acceptable, depending on the task).

We train several variations of pooling classifiers, as shown in Table \ref{results table}. First, we train classifiers end-to-end on the real/fake task, varying the style of word embedding. The classifier portion consists only of a single softmax layer. We evaluate these classifiers on CoLA without CoLA training. 

Second, we train pooling classifiers entirely on CoLA. We test only ELMo-style embeddings here because, unlike BNC and GloVe embeddings, they include robust contextual information about the entire sequence, eliminating the need for training a large LSTM on CoLA alone. 

Third, we transfer features learned from the real/fake task to classifiers trained on CoLA. Specifically, we freeze the weights of the sentence encoder portion of the real/fake classifiers, and train new classifiers on CoLA using the sentence embeddings as input. For these experiments, in addition to a sigmoid layer, the classifier has an additional hidden $tanh$ layer to compensates for the fact that the sentence encoder is not fine-tuned on CoLA.

\paragraph{Lau et al.~(2016)}

We compare our models to those of \newcite{lau2016cognitive}. Their models obtain an acceptability prediction from unsupervised LMs by normalizing the LM output using one of several metrics. Following their recommendation, we use the \textit{Word LogProb Min-1} metric.\footnote{Where $s$=sentence, p$_{\textnormal{LM}}(x)$ is the probability the LM assigns to string $x$ and p$_{\textnormal{u}}(x)$ is the unigram probability of string $x$: Word LP Min-1 = min$\left\{-\frac{\textnormal{log p}_{\textnormal{LM}}(w)}{\textnormal{log p}_{\textnormal{u}}(w)},\ w \in s\right\}$. Lau et al.\ also get strong results with the \emph{SLOR} metric. We also calculate results with \emph{SLOR} but find them to be slightly worse overall, though not universally. We do not report these results, but they are available upon request.} Since this metric produces unbounded scalar scores rather than probabilities or Boolean judgments, we fit a threshold to the outputs in order to use these models as acceptability classifiers. This is done with 10-fold cross-validation on the CoLA test set: We repeatedly find the optimum threshold for 90\% of the model outputs and evaluate the remaining 10\% with that threshold, until all the data have been evaluated.
Following their methods, we train $n$-gram models on the BNC using their published code.\footnote{\url{https://github.com/jhlau/acceptability\_prediction}}
In place of their RNN LM, we use the same LSTM LM that we use to generate sentences for the real/fake task.

\paragraph{CBOW}

For a simple baseline, we train a continuous bag-of-words (CBOW) model directly on CoLA. We pass the sum of BNC word embeddings for the sentence to a multilayer perceptron with a single hidden layer.

\subsection{Training details}\label{sec:training}

All neural network models are implemented in PyTorch and optimized using Adam \cite{kingma2014adam}.
We train 20 LSTM LMs with from-scratch embeddings for up to 7 days or until completing 4 epochs without improving in development perplexity and select the best checkpoint. Hyperparameters for each experiment are chosen at random in these ranges: embedding size $\in$ [200, 600], hidden size $\in$ [600, 1200], number of layers $\in$ [1, 4], learning rate $\in$ [$3\times 10^{-3}$, $10^{-5}$], dropout rate $\in$ \{0.2, 0.5\}. We select the model with best performance for use in further experiments.

We train 20 pooling classifiers end-to-end on real/fake data with BNC embeddings, 20 with GloVe, and 20 with ELMo-style embeddings for up to 7 days or until completing 4 epochs without improving in development MCC. We train 20 pooling classifiers end-to-end on CoLA using ELMo-style embeddings. Hyperparameters are chosen at random in these ranges: embedding size $\in$ [200, 600], hidden size $\in$ [500, 1500], number of layers $\in$ [1, 5], learning rate $\in$ [$3\times 10^{-3}$, $10^{-5}$], dropout rate $\in$ \{0.2, 0.5\}.

For transfer learning experiments, we extract and freeze the weights from the encoders from the 5 best real/fake classifiers with BNC, GloVe, and ELMo-style embeddings, each. For every encoder, we train 10 classifiers on CoLA until completing 20 epochs without improving in MCC on the development set. Hyperparameters are chosen at random in these ranges: hidden size $\in$ [20, 1200] and learning rate $\in$ [$10^{-2}$, $10^{-5}$], dropout rate $\in$ \{0.2, 0.5\}.

For our single best model---a pooling classifier with ELMo-style embeddings, an encoder with real/fake training, and a classifier with CoLA training---the embedding size (i.e.~LM hidden size) is 819 dimensions, the real/fake encoder hidden layer size is 528 dimensions, and the acceptability classifier hidden layer size is 1134.

\section{Results and Discussion}\label{results}

Table \ref{results table} shows the results of the best run from each experiment. The best model overall is the real/fake model with ELMo-style embeddings. It achieves the highest MCC and accuracy both in-domain and out-of-domain by a large margin, outperforming even the models with access to GloVe.

All models with real/fake encoders and CoLA training perform better than the unsupervised models of \newcite{lau2016cognitive} on both evaluation metrics on the in-domain test set. Out-of-domain, Lau et al.'s baselines offer the second-best results. Our models consistently perform worse out-of-domain than in-domain, with MCC dropping by as much as 50\% in one case. Since \citeauthor{lau2016cognitive}'s baselines don't use the training set, they perform similarly in-domain and out-of-domain. Real/fake classifiers without any additional training on CoLA tend to perform significantly worse than their counterparts with CoLA supervision.

The sequence models consistently outperform the word order-independent CBOW baseline, indicating that the LSTM models are using word order for acceptability classification in a non-trivial way. In line with Lau et al.'s findings, the $n$-gram LM baselines are worse than the LSTM LM. This result is expected given that $n$-gram models, but not LSTMs, have a limited feature window.

\paragraph{Discussion}

Of the models we have tested, LSTMs are the most effective low-bias learners for acceptability classification. Compared to humans, though, their absolute performance is underwhelming. This indicates to us that while the ANNs we study can acquire substantial knowledge of grammar, their linguistic competence is far from rivaling humans'.

Our models with unsupervised pretraining have an advantage over similar models without pretraining. This finding aligns with the conclusions of \citet{peters2018elmo}. We see this effect with both the LM pretraining for our ELMo-style embeddings real/fake pretraining for our sentence encoders. Unsurprisingly, the unsupervised \citeauthor{lau2016cognitive} models and real/fake classifiers are not as effective as models trained on CoLA. However, they far outperform random guessing and the CBOW baseline, indicating that even purely unsupervised models acquire significant knowledge of grammar.

The supervised models universally see a substantial drop in performance from the in-domain test set to the out-of-domain test set. This suggests that they have specialized somewhat to the phenomena in the training set, rather than English grammar in a fully general way as one would hope for. Addressing this problem will likely involve new forms of regularization to mitigate this overfitting and, more importantly, new pretraining strategies that can help the model better learn the fundamental ingredients of grammaticality from unlabeled data.

\section{CoLA Design Experiments}\label{sec:cola design}

The results in the previous section highlight the effects of pretraining, but give little insight into how the labeled training data in CoLA impacts classifier performance. To quantify the impact of CoLA training, we conduct two additional experiments: First, we measure how the amount of training data impacts model performance on the CoLA development set. Second, we investigate how the specific contents of the in-domain and out-of-domain sets impact model generalization. 

\paragraph{Training Set Size}

\begin{figure}
    \centering
    \includegraphics[width=0.5\textwidth]{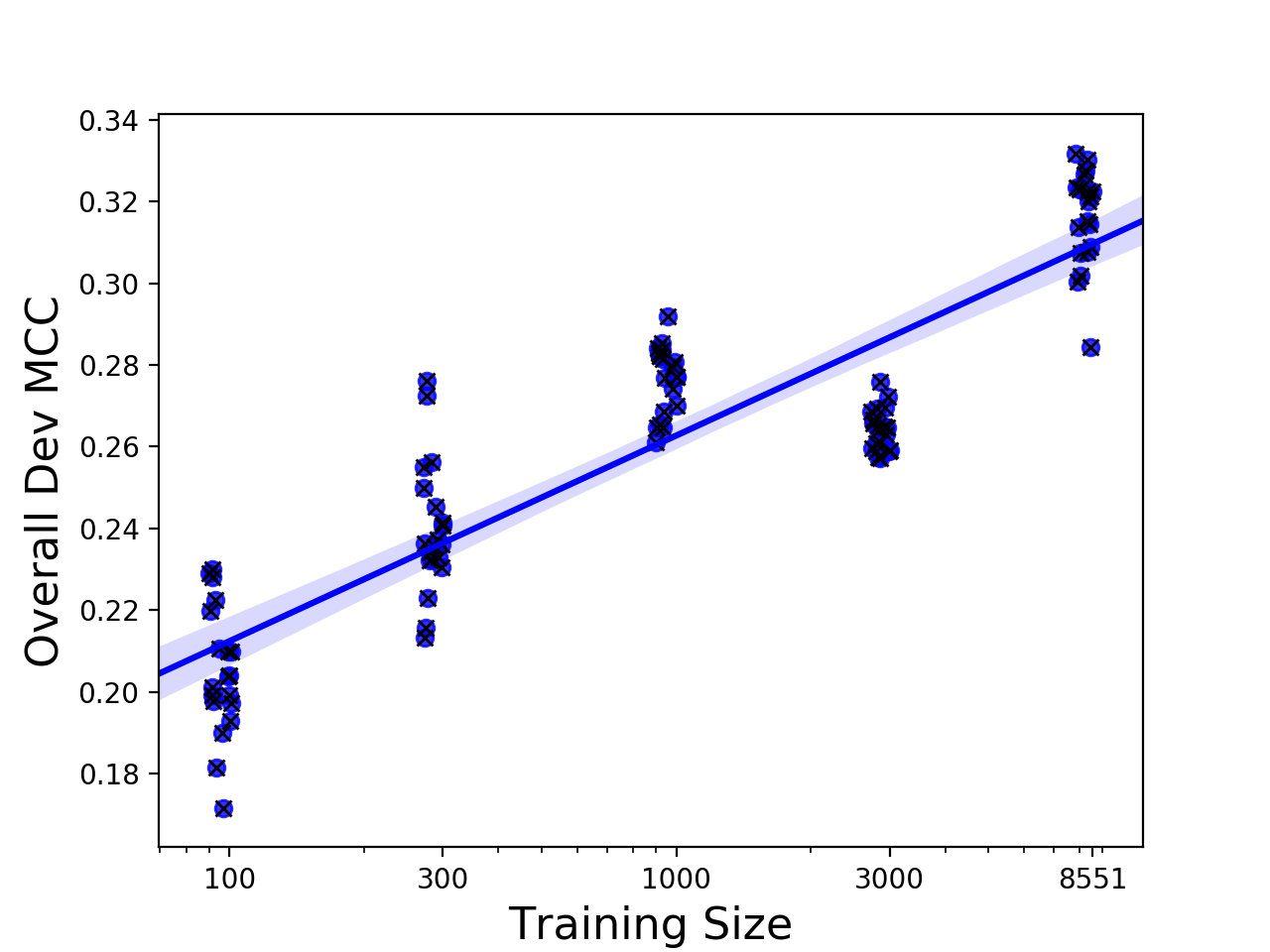}
    \caption{Results on the CoLA development set as a function of the number of training examples, with line of best fit and 95\% confidence interval. Random $x$ jitter added.}
    \label{fig:sizes}
\end{figure}

In this experiment, we vary the amount of training data seen by our acceptability classifiers. We construct alternate training sets of sizes 100, 300, 1000, and 3000 by randomly downsampling the 8551-example CoLA training set. Then, for each training set we train classifiers with 20 restarts using the best performing ELMo-style real/fake encoder, and evaluate on the entire development set. Figure \ref{fig:sizes} plots the results. As training data increases from 100 to 8551 sentences, we see approximately log-linear improvements in performance. The small decrease in performance between 1000 and 3000 sentences is likely an artifact of the random downsampling.

From these results we draw two main conclusions: First, it appears that increasing the amount of training data in CoLA by an order of magnitude may significantly benefit our models. Second, much of what our models learn from CoLA can be learned from as few as 300 training examples. This suggests that CoLA training is not teaching our models specific facts about acceptability as much as teaching them to use existing grammatical knowledge from the sentence encoders.

\paragraph{Splitting CoLA}

\begin{table*}[h]
    \centering \small
    \begin{tabular}{c c c c c c c p{0.35\textwidth} r}
    \toprule
\multirow{2}{*}{\textbf{Split}}&
\multicolumn{2}{c}{\textbf{In-Domain}}&
\multicolumn{2}{c}{\textbf{Out-of-Domain}}&
\multicolumn{2}{c}{\textbf{Overall}}&
\textbf{Out Sources}&
\textbf{Out N}\\
& Acc. & MCC & Acc. & MCC & Acc. & MCC \\\midrule

orig. & 
0.701&
0.348&
0.620&
0.223&
0.660&
0.285&
C05, J71, S97, CLC95, W80, SWB04&
1049
\\

1&
0.729&
0.357&				
0.632&	
0.195&		
0.680&
0.275&	
BC01, B73&
1139
\\

2&
0.700&
0.319&
0.666&
0.188&
0.683&
0.255&
KL93, SGWW85, W80, D98, B73, G81&
853\\

3&
0.708&
0.333&
0.659&
0.284&
0.684&
0.307&	
AD03, D98, G81&
1237\\

4&
0.663&
0.243&
0.673&
0.267&
0.668&
0.252&
B82, SWB04, CJ99&
789\\

5&
0.720&
0.349&
0.671&
0.285&
0.696&
0.315&	
M02, BC01, CJ99&
1539\\

    \bottomrule
    \end{tabular}
    \caption{Results for 5 different splits of CoLA and the original split into in-domain and out-of-domain. All results are averages over 20 restarts. Out N is the number of out-of-domain sentences. Sources are abbreviated by authors' last initial and year; full citations for each source are shown in Table \ref{sources}.}
    \label{tab:splits}
\end{table*}

Our results in Table \ref{results table} show that our models' performance drops noticeably when tested on out-of-domain sentences from publications not represented in the training data. In this experiment, we investigate different splits of CoLA into in-domain and out-of-domain to test the degree to which the decrease in performance on out-of-domain sentences is a stable property of these models, or simply an artifact of the particular publications represented in the out-of-domain set (as described in section \ref{cola}). 

The splits are constructed by randomly selecting sources from the 23 sources from CoLA to hold out until the sum of their sizes exceeds 750. This gives out-of-domain set sizes ranging from 789 to 1539, consisting of 2 to 6 sources. CoLA's original out-of-domain set contains 1049 examples and 6 sources. Development and test sets are constructed by randomly splitting the out-of-domain data in half, and randomly selecting an approximately equal number of in-domain sentences. For each training set we train classifiers with 20 restarts using the encoder from the best performing ELMo-style real/fake classifier.

In Table \ref{tab:splits}, we report the average test performance over 20 restarts. We conclude that the domain difference between two samples of sources in CoLA is generally a meaningful one for these models. This is especially so for the original split, where average in-domain MCC is 0.125 greater than out-of-domain MCC, close to the maximum observed difference of 0.162. By contrast, in one case average out-of-domain performance was actually better. This tells us that the particular nature of the sources in each domain has a large effect on what our models learn.

%We also observe that larger out-of-domain size correlates with better overall performance. Since we evaluate the checkpoint of our models that performs best on the development set, having a larger out-of-domain set (which implies a larger development set) decreases the chance of selecting a checkpoint which overfits the development set data.

\section{Phenomenon-Specific Analysis}\label{analysis}

In addition to testing the general grammatical knowledge of low-bias learners, acceptability classification can be used to probe models' knowledge of particular linguistic phenomena. We analyze our baselines' performance by phenomenon using two methods: First, we break down their performance on CoLA based on the different constructions present in the target sentences. Second, we evaluate them on controlled test sets targeting specific grammatical contrasts.

\subsection{CoLA Performance by Phenomenon}

\begin{figure}
    \centering
    \includegraphics[width=0.5\textwidth]{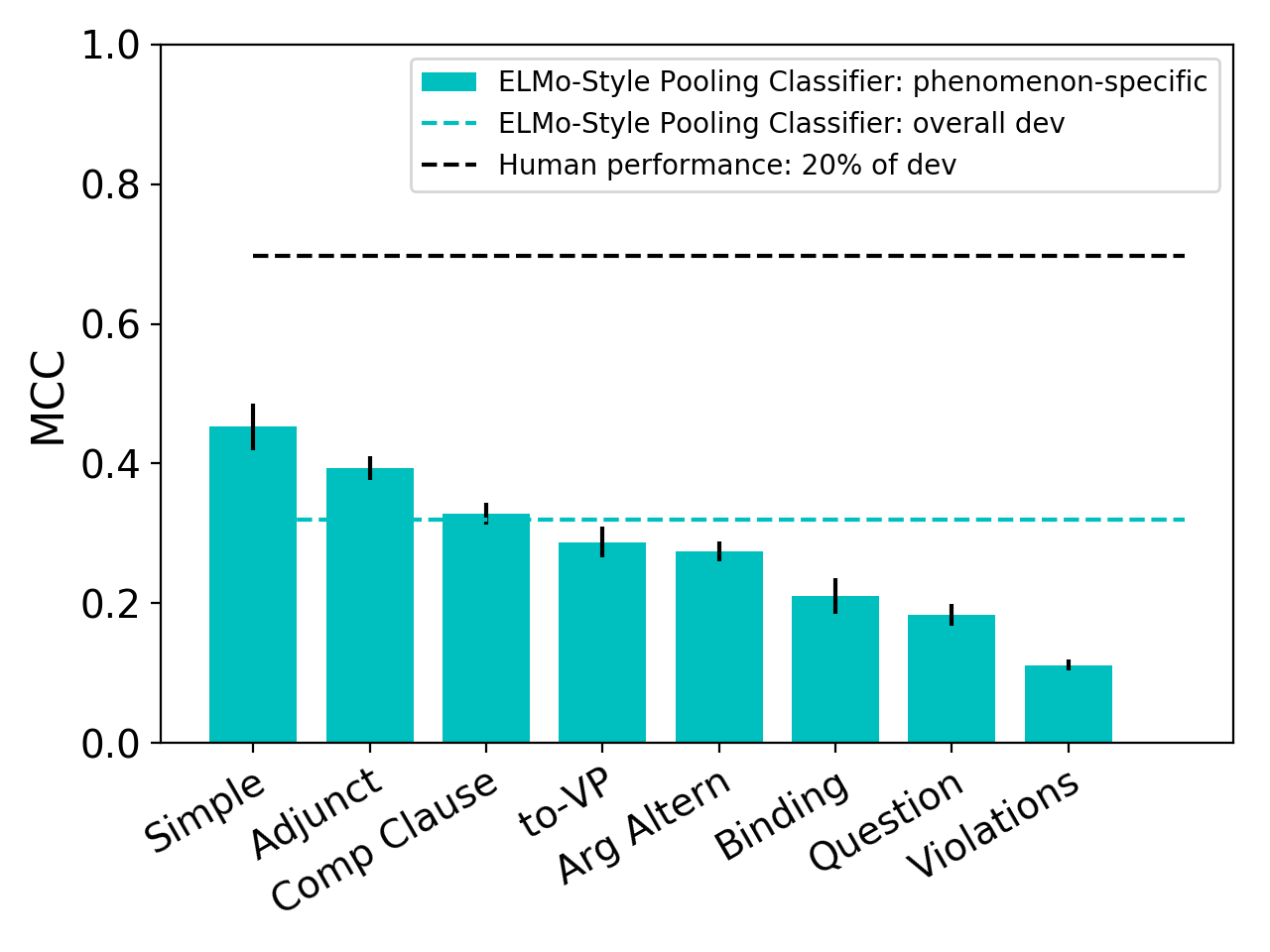}
    \caption{Performance on phenomenon-specific subsets of the CoLA development set. Results are the mean over 20 random restarts, with error bars $\pm 1$ STD. Lines show mean performance on the entire dev set and mean human performance on 200 dev set sentences. \emph{Simple}: no marked syntactic structures. \emph{Adjunct}: adjuncts of nouns and verb phrases. \emph{Comp Clause}: embedded or complement clauses. \emph{to-VP}: non-finite embedded verb phrase. \emph{Binding}: pronouns and binding phenomena. \emph{Question}: questions and relative clauses. \emph{Violations}: morphological or semantic violations, or an extra/missing word. }
    \label{fig:categories}
\end{figure}

In this error analysis, we study performance on CoLA as a function of the syntactic features of the individual sentences, using the 8 features described in Section \ref{cola}. We train classifiers with 20 restarts using the best performing ELMo-style real/fake encoder. For each feature, we measure the MCC of our models on only those sentences with that feature. 

Figure \ref{fig:categories} shows the mean MCC over 20 restarts for each feature. Unsurprisingly, syntactically simple sentences are easier than average, but unexpectedly sentences with adjuncts are as well. Sentences with complement clauses, embedded VPs, and argument alternations are about as hard as the average sentence in CoLA. While these constructions can be complex, they also occur with very high frequency. Sentences with binding and violations, including morphological violations, are among the hardest. We also find that our models perform poorly on sentences with question-like syntax. This difficulty is likely due to long-distance dependencies in these sentences.

\subsection{Targeted Test Sets}

Here, we run additional evaluations to probe whether our models can reliably classify sets of sentences which target a single grammatical contrast. This kind of evaluation can give insight into what kinds of grammatical features our models do and do not acquire easily. Using data generation techniques inspired by \newcite{ettinger2016probing}, we build five auxiliary datasets (described below) using simple rewrite grammars which target specific grammatical contrasts.
% Specifically, we investigate the models' understanding of gross word order (``SVO'' in Table \ref{artificial}), long-distance dependencies between wh-words and gaps (``wh-extraction''), verbal argument structure alternations (``inchoative''), number agreement (``singular/pl''), and anaphoric dependency (``reflexive'').

Unlike in CoLA, none of these judgments are meant to be difficult or subtle, and we expect that most humans could reach perfect accuracy. We also take care to make the test sentences as simple as possible to reduce classification errors unrelated to the target contrast. Specifically, we limit noun phrases to 1 or 2 words and use semantically related vocabulary items within examples.

\paragraph{Subject-Verb-Object} This test set consists of 100 triples of subject, verb, and object each appearing in five permutations of (SVO, SOV, VSO, VOS, OVS).\footnote{OSV is excluded because it does not yield a clear acceptability rating. Examples such as ``The book John read'', can be interpreted as marginally acceptable sentences with topicalized subjects, or as acceptable noun phrases (rather than sentences) with relative clause modifiers.} The set of 100 triples is the Cartesian product product of three sets containing 10 subjects (\{John, Bo, ...\}), 2 verbs (\{read, wrote\}), and 5 objects (\{the book, the letter, ...\}).

\ex. a. \phantom{*}Bo read the book. \hfill b. *Bo the book read. \\c. *read Bo the book. \hfill d. *read the book Bo. \\e. *the book read Bo.

\begin{table*}[h]
\centering\small
\begin{tabular}{llllrrrrr }
\toprule
\textbf{Model}&
\textbf{Emb.}&
\textbf{Enc.}&
\textbf{Class.}&
\textbf{SVO}&
\textbf{Wh}&
\textbf{Causative}&
\textbf{SV Agr.}&
\textbf{Reflexive}\\\midrule
%
% LSTM LM SLOR&
% BNC&
% --&
% CoLA Th.&
% 0.924&
% 0.063&
% 0.283&
% 0.323&
% \textbf{0.521}\\
%
LSTM LM WLPM&
BNC&
--&
CoLA Th.&
0.801&
\textbf{0.601}&
0.270&
\textbf{0.599}&
\textbf{0.152}\\\midrule
Pooling&
ELMo-St.&
CoLA&
CoLA&
0.637&
0.102&
\textbf{0.633}&
0.128&
0.075\\
Pooling&
BNC&
R/F&
CoLA&
0.381&
0.184&
0.463&
0.098&
0.043\tabularnewline
Pooling&
GloVe&
R/F&
CoLA&
\textbf{0.988}&
0.059&
0.614&
0.277&
0.150\\
Pooling&
ELMo-St.&
R/F&
CoLA&
0.650&
0.000&
0.449&
0.302&
-0.020\tabularnewline
\bottomrule
\end{tabular}
\caption{MCC results for specific phenomena. \emph{Emb}.~is model embedding style; \emph{Enc}.~is model encoder training, \emph{Class}.~is model classifier training. \emph{R/F} is real/fake, \emph{ELMo-St.} is ELMo-style, and \emph{CoLA-Th.} is threshold tuning on CoLA. \emph{LSTM LM SLOR/WLPM} is the LM with Lau et al. metrics Word LP Min-1. 
}\label{artificial}
\end{table*}

\paragraph{Wh-Extraction} This test set consists of 260 pairs of contrasting examples, as in \ref{wh}. This is to test (i) whether a model has learned that a wh-word must correspond to a gap in the sentence, and (ii) whether the model can identify non-local dependencies up to three words away. The data contain 10 first names as subjects and 8 sets of verbs and related objects \ref{set-2}. Every compatible verb-object pair appears with every subject. 

\ex.\label{wh}
\a. What did John fry?
\b. *What did John fry the potato?

\ex.\label{set-2} \{\{boil, fry\}, \{the egg, the potato\}\}

\paragraph{Causative-Inchoative Alternation} This test set is based on a syntactic alternation conditioned by the lexical semantics of particular verbs. It contrasts verbs like \emph{popped} which undergo the causative-inchoative alternation, with verbs like \emph{blew} that do not. If \emph{popped} is used transitively \ref{cause}, the subject (\emph{Kelly}) is an agent who causes the direct object (\emph{the bubble}) to change states. Used intransitively \ref{inch}, it is the subject (\emph{the bubble}) that undergoes a change of state and the cause need not be specified \cite{levin1993verb}. The test set includes 91 verb/object pairs, and each pair occurs in the two forms as in \ref{cause/inch}. 36 pairs allow the alternation, and the remaining 55 do not.

\ex.\label{cause/inch}
\a.\label{cause}Kelly popped/blew the bubble.
\b.\label{inch}The bubble popped/*blew.

\paragraph{Subject-Verb Agreement} This test set is generated from 13 subjects in singular and plural form crossed with 13 verbs in singular and plural form. This gives 169 quadruples as in \ref{agr}.

\ex.\label{agr}
\a.My friend has/*have to go.
\b.My friends *has/have to go.

\paragraph{Reflexive-Antecedent Agreement} This test set probes whether a model has learned that every reflexive pronouns must agree with an antecedent noun phrase in person, number, and gender. The dataset consists of a set of 4 verbs crossed with 6 subject pronouns and 6 reflexive pronouns, giving 144 sentences, only 1 out of 6 acceptable. 

\ex.\label{refl} I amused myself/*yourself/*herself/*himself/ *ourselves/*themselves.

\paragraph{Results}

The results from these experiments are given in Table \ref{artificial}. Our models' performance on these test sets is mixed. They make some systematic acceptability judgments that reflect correct grammatical generalizations. Some models are very effective at judging violations in gross word order (\textit{SVO} in Table \ref{artificial}). The pooling classifier with GloVe embeddings achieves near perfect correlation, suggesting that it systematically uses gross word order. However, the remaining tests yield much poorer performance.

% TODO: Building on what you say about the messy results by domain, comment on the fact that there isn't a clean correlation across phenomena in which model does best.

Our models consistently outperform Lau et al.'s baselines on lexical semantics (\textit{Causative}), judging more accurately whether a verb can undergo the causative-inchoative alternation. This may be due in part to the fact that our models receive supervision from CoLA, in which argument alternations are well represented (see Figure \ref{fig:freqs}).

Lau et al.'s baseline outperforms our models in some cases. The LSTM LM with the Word LP Min-1 metric is the only model that can reliably identify the non-local dependency between a \textit{wh}-word and its gap (\textit{Wh-extraction}). It also performs relatively better on judgments involving agreement (\textit{Singular/Pl}). All models struggle on the \textit{Reflexive} examples. 

The poor performance of our models on contrasts involving agreement (\textit{Singular/Pl} and \textit{Reflexive}) is surprising in light of findings by \cite{linzen2016assessing} that LSTMs can identify agreement errors easily even without access to sub-word information. We speculate that this is due to under-representation of the relevant examples in CoLA. We estimate that morphological violations make up about 6\% of examples in CoLA (about half of the \emph{Violations} in Figure \ref{fig:freqs}).

\section{Motivation \& Related Work}\label{sec:discussion}

% \subsection{Linguistic Competence of ANNs}

% Recent NLU systems based on ANNs seem to be using language in increasingly human-like ways. This undeniable trend leads us to ask whether ANNs also acquire knowledge of grammar comparable to that of humans. We have presented several experiments which apply the acceptability classification task to this end. 
We see two chief motivations that guide work on acceptability classification with ANNs by us and by others: First, more fine-grained evaluation tools may accelerate work on general-purpose neural network modules for sentence understanding. Second, studying the linguistic competence of ANNs bears on foundational questions in linguistics about the learnability of grammar. 

\paragraph{Fine-Grained Evaluation of ANNs}

The question of how well ANNs learn fine-grained grammatical distinctions has been the subject of much recent work. One method is to train models to perform probing tasks which target a construction of interest. Examples of such tasks are to determine whether the sentence is in active or passive voice \cite{shi2016syntax}, whether the subject is singular or plural \cite{conneau2018cram}, or whether a given token is under the scope of negation \cite{ettinger2018assessing}. In each case, the authors use these tasks to compare the performance of reusable sentence embeddings. 

Acceptability classification can be used to target many of the same grammatical constructions as probing tasks. For instance, an acceptability classifier that can reliably distinguish between pairs of sentences as in \ref{eg:reproduce probing} must have implicit knowledge of the whether the subject of a sentence is singular or plural (in the first case) and whether the token \emph{ever} is under the scope of negation. These exact experiments have been conducted by \newcite{linzen2016assessing} and \newcite{marvin2018targeted}, respectively, although these works differ from our approach in that they do not evaluate domain general acceptability classifiers on these contrasts.

\ex. \label{eg:reproduce probing}
\a. The key is/*are on the table.
\b. Betsy hasn't/*has ever been to France.

\noindent

Acceptability classification also enables certain kinds of investigations not possible with probing tasks. A single acceptability classifier can be trained to identify numerous unrelated contrasts. This is generally not possible with probing tasks, because the classes are tied to specific grammatical concepts. Acceptability classification also encourages direct comparison between ANN and human linguistic competence because, unlike many probing tasks, it can be easily performed by native speakers without linguistic training. Finally acceptability classifiers and generative grammars share a common objective, namely to predict the well-formedness of all and only those strings of the language that are acceptable to native speakers. Accordingly, it is straightforward to draw parallels between acceptability classifiers and established work in generative linguistics.

\paragraph{The Poverty of the Stimulus}

Research on acceptability classification can also be brought to bear on a foundational question in linguistic theory: The extent to which human linguistic competence is learned or innate. The influential \emph{argument from the poverty of the stimulus} (APS) holds that the extent of human linguistic competence cannot be explained by purely domain general learning mechanisms and that humans must be born with a Universal Grammar which imparts specific knowledge of grammatical universals to the child and makes learning possible \cite{chomsky1965aspects}. While the APS has been subject to much criticism \cite{pullum2002empirical}, it remains a foundation of much of contemporary linguistics.

In the setting of machine learning, the APS predicts that any artificial leaner trained with no prior knowledge of the principles of syntax and no more data than a human child sees must fail to make acceptability judgments with human-level accuracy \cite{clark2011nativism}. If linguistically-uninformed neural network models achieve human-level performance on specific phenomena or on a domain-general dataset like CoLA, this would be clear evidence limiting the scope of phenomena for which the APS can hold. 

However, acceptability classification alone cannot evaluate aspects of ANNs' linguistic competence against humans' in every relevant way. For example, \newcite{berwick2011poverty} note that native speakers can easily recognize that, e.g., in \emph{Bo is easy to please}, Bo is the entity being \emph{pleased}, while in \emph{Bo is eager to please}, Bo is the one who does the \emph{pleasing}. Since the acceptability judgments in CoLA are reading-independent (see Table \ref{include exclude}), they cannot be used to probe whether ANNs understand these distinctions.

We wish to stress that the success of supervised acceptability classifiers like the ones we train cannot falsify the APS, because unacceptable examples play no apparent role in child language acquisition. While unsupervised acceptability classification could do so, more work is needed to find methods for extracting reliable Boolean acceptability judgments from unsupervised language models. Our approach of fitting a threshold to the models of \newcite{lau2016cognitive} gives encouraging results, but these models are ultimately not as effective as supervised models. An alternative adopted by \newcite{linzen2016assessing} and \newcite{marvin2018targeted} is to evaluate whether language models' assign higher probability to the acceptable sentence in a minimal pair. However, this forced choice minimal pair task, as discussed in Section \ref{sec:minimal pairs}, cannot be applied to CoLA, which does not exclusively contain minimal pairs.

Still, we maintain that our approach is a valuable step in the direction of evaluating the APS. Our results strongly suggest that grammatically unbiased sentence embeddings and contextualized word embeddings have non-trivial implicit knowledge of grammar before supervised training on CoLA. As our experiments in Section \ref{sec:cola design} show, a significant portion of what these models learn from CoLA can be learned from relatively little acceptability judgment data (as few as 300 sentences, of which fewer than 100 are unacceptable). Furthermore, the real/fake encoders and ELMo-style embeddings are trained on a quantity of data comparable to what human learners are exposed to. Given the rapid pace of development of new robust sentence embeddings, we expect to see increasingly human-like acceptability judgments from powerful neural networks in coming years, though with an eye towards evaluating the APS, future work should continue to investigate acceptability classifiers with unsupervised methods and restricted training resources.

\section{Conclusion}

% TODO: Mention GLUE in final version.

This work offers resources and baselines for the study of semi-supervised machine learning for acceptability judgments. 
Most centrally, we introduce CoLA, the first large-scale corpus of acceptability judgments, making it possible to train and evaluate modern neural networks on this task. In baseline experiments, we find that a network trained on our artificial real/fake task, combined with ELMo-style word representations, outperforms other available models, but remains far from human performance.

Much work remains to be done to implement the agenda described in Section \ref{sec:discussion}. There is much untapped potential in the acceptability classification task as a fine-grained evaluation tool and as a test of the Poverty of the Stimulus Argument. We hope for future work to test the performance of a broader range of new effective low-bias machine learning models on CoLA, and to investigate further what grammatical principles these models do and do not learn.

% TODO: Go through all the arXiv papers you cite and see if there's a permanent/published version in ACL/NIPS/ICLR/etc.. I know that there is for most of them, and you're expected to cite that when possible.

\section*{Acknowledgments}

This project has benefited from help and feedback
at various stages from Chris Barker, Pablo Gonzalez, Shalom Lappin, Omer Levy, Marie-Catherine
de Marneffe, Alex Wang, Alexander Clark, everyone in the
Deep Learning in Semantics seminar at NYU, and three anonymous TACL reviewers. This
project has benefited from financial support to SB by
Google, Tencent Holdings, and Samsung Research. This material is based upon work supported by the National Science Foundation under Grant No. 1850208. Any opinions, findings, and conclusions or recommendations expressed in this material are those of the author(s) and do not necessarily reflect the views of the National Science Foundation.

\bibliography{tacl2018}

\begin{thebibliography}{59}
\expandafter\ifx\csname natexlab\endcsname\relax\def\natexlab#1{#1}\fi

\bibitem[{Adger(2003)}]{adger2003core}
David Adger. 2003.
\newblock \emph{Core Syntax: A Minimalist Approach}.
\newblock Oxford University Press Oxford.

\bibitem[{Adi et~al.(2017)Adi, Kermany, Belinkov, Lavi, and
  Goldberg}]{adi2017fine}
Yossi Adi, Einat Kermany, Yonatan Belinkov, Ofer Lavi, and Yoav Goldberg. 2017.
\newblock Fine-grained analysis of sentence embeddings using auxiliary
  prediction tasks.
\newblock In \emph{Proceedings of ICLR Conference Track. Toulon, France.}

\bibitem[{Baltin(1982)}]{baltin1982landing}
Mark~R. Baltin. 1982.
\newblock A landing site theory of movement rules.
\newblock \emph{Linguistic Inquiry}, 13(1):1--38.

\bibitem[{Baltin and Collins(2001)}]{baltin2001handbook}
Mark~R. Baltin and Chris Collins, editors. 2001.
\newblock \href {https://doi.org/10.1111/b.9781405102537.2003.x}
  {\emph{Handbook of Contemporary Syntactic Theory}}.
\newblock Blackwell Publishing Ltd.

\bibitem[{Berwick et~al.(2011)Berwick, Pietroski, Yankama, and
  Chomsky}]{berwick2011poverty}
Robert~C Berwick, Paul Pietroski, Beracah Yankama, and Noam Chomsky. 2011.
\newblock Poverty of the stimulus revisited.
\newblock \emph{Cognitive Science}, 35(7):1207--1242.

\bibitem[{Bird and Loper(2004)}]{bird2004nltk}
Steven Bird and Edward Loper. 2004.
\newblock {NLTK}: the natural language toolkit.
\newblock In \emph{Proceedings of the ACL 2004 on Interactive poster and
  demonstration sessions}, page~31. Association for Computational Linguistics.

\bibitem[{Bresnan(1973)}]{bresnan1973comparative}
Joan~W. Bresnan. 1973.
\newblock Syntax of the comparative clause construction in {E}nglish.
\newblock \emph{Linguistic Inquiry}, 4(3):275--343.

\bibitem[{Carnie(2013)}]{carnie2013syntax}
Andrew Carnie. 2013.
\newblock \emph{Syntax: A Generative Introduction}.
\newblock John Wiley \& Sons.

\bibitem[{Chomsky(1957)}]{chomsky1957syntactic}
Noam Chomsky. 1957.
\newblock \emph{Syntactic Structures.}
\newblock Mouton.

\bibitem[{Chomsky(1965)}]{chomsky1965aspects}
Noam Chomsky. 1965.
\newblock \emph{Aspects of the Theory of Syntax}.
\newblock MIT Press.

\bibitem[{Chomsky(1995)}]{chomsky1995minimalist}
Noam Chomsky. 1995.
\newblock \emph{The {M}inimalist {P}rogram}.
\newblock MIT press.

\bibitem[{Chung et~al.(1995)Chung, Ladusaw, and McCloskey}]{chung1995sluicing}
Sandra Chung, William~A. Ladusaw, and James McCloskey. 1995.
\newblock Sluicing and logical form.
\newblock \emph{Natural Language Semantics}, 3(3):239--282.

\bibitem[{Clark and Lappin(2011)}]{clark2011nativism}
Alexander Clark and Shalom Lappin. 2011.
\newblock \emph{Linguistic Nativism and the Poverty of the Stimulus}.
\newblock John Wiley \& Sons.

\bibitem[{Collins(2005)}]{collins2005smuggling}
Chris Collins. 2005.
\newblock A smuggling approach to the passive in {E}nglish.
\newblock \emph{Syntax}, 8(2):81--120.

\bibitem[{Conneau et~al.(2017)Conneau, Kiela, Schwenk, Barrault, and
  Bordes}]{conneau2017supervised}
Alexis Conneau, Douwe Kiela, Holger Schwenk, Lo{\"\i}c Barrault, and Antoine
  Bordes. 2017.
\newblock Supervised learning of universal sentence representations from
  natural language inference data.
\newblock In \emph{Proceedings of the 2017 Conference on Empirical Methods in
  Natural Language Processing}, pages 670--680.

\bibitem[{Conneau et~al.(2018)Conneau, Kruszewski, Lample, Barrault, and
  Baroni}]{conneau2018cram}
Alexis Conneau, German Kruszewski, Guillaume Lample, Lo{\"\i}c Barrault, and
  Marco Baroni. 2018.
\newblock What you can cram into a single \&!\#* vector: Probing sentence
  embeddings for linguistic properties.
\newblock In \emph{ACL 2018-56th Annual Meeting of the Association for
  Computational Linguistics}, volume~1, pages 2126--2136. Association for
  Computational Linguistics.

\bibitem[{Culicover and Jackendoff(1999)}]{culicover1999comparative}
Peter~W. Culicover and Ray Jackendoff. 1999.
\newblock The view from the periphery: The {E}nglish comparative correlative.
\newblock \emph{Linguistic Inquiry}, 30(4):543--571.

\bibitem[{Dayal(1998)}]{dayal1998inherently}
Veneeta Dayal. 1998.
\newblock Any as inherently modal.
\newblock \emph{Linguistics and Philosophy}, 21(5):433--476.

\bibitem[{Ettinger et~al.(2018)Ettinger, Elgohary, Phillips, and
  Resnik}]{ettinger2018assessing}
Allyson Ettinger, Ahmed Elgohary, Colin Phillips, and Philip Resnik. 2018.
\newblock Assessing composition in sentence vector representations.
\newblock In \emph{Proceedings of the 27th International Conference on
  Computational Linguistics}, pages 1790--1801. Association for Computational
  Linguistics.

\bibitem[{Ettinger et~al.(2016)Ettinger, Elgohary, and
  Resnik}]{ettinger2016probing}
Allyson Ettinger, Ahmed Elgohary, and Philip Resnik. 2016.
\newblock Probing for semantic evidence of composition by means of simple
  classification tasks.
\newblock In \emph{Proceedings of the 1st Workshop on Evaluating Vector-Space
  Representations for NLP}, pages 134--139.

\bibitem[{Futrell et~al.(2018)Futrell, Wilcox, Morita, and
  Levy}]{futrell2018rnns}
Richard Futrell, Ethan Wilcox, Takashi Morita, and Roger Levy. 2018.
\newblock {RNN}s as psycholinguistic subjects: Syntactic state and grammatical
  dependency.
\newblock \emph{arXiv preprint arXiv:1809.01329}.

\bibitem[{Gazdar(1981)}]{gazdar1981unbounded}
Gerald Gazdar. 1981.
\newblock Unbounded dependencies and coordinate structure.
\newblock In \emph{The Formal Complexity of Natural Language}, pages 183--226.
  Springer.

\bibitem[{Gibson and Fedorenko(2010)}]{gibson2010quantitative}
Edward Gibson and Evelina Fedorenko. 2010.
\newblock Weak quantitative standards in linguistics research.
\newblock \emph{Trends in Cognitive Sciences}, 14(6):233--234.

\bibitem[{Goldberg and Jackendoff(2004)}]{goldberg2004resultative}
Adele~E. Goldberg and Ray Jackendoff. 2004.
\newblock The {E}nglish resultative as a family of constructions.
\newblock \emph{Language}, 80(3):532--568.

\bibitem[{Hart and Risley(1992)}]{hart1992parenting}
Betty Hart and Todd~R. Risley. 1992.
\newblock American parenting of language-learning children: Persisting
  differences in family-child interactions observed in natural home
  environments.
\newblock \emph{Developmental Psychology}, 28(6):1096.

\bibitem[{Heilman et~al.(2014)Heilman, Cahill, Madnani, Lopez, Mulholland, and
  Tetreault}]{heilman2014grammaticality}
Michael Heilman, Aoife Cahill, Nitin Madnani, Melissa Lopez, Matthew
  Mulholland, and Joel Tetreault. 2014.
\newblock Predicting grammaticality on an ordinal scale.
\newblock In \emph{Proceedings of the 52nd Annual Meeting of the Association
  for Computational Linguistics (Volume 2: Short Papers)}, volume~2, pages
  174--180.

\bibitem[{Jackendoff(1971)}]{jackendoff1971gapping}
Ray~S. Jackendoff. 1971.
\newblock Gapping and related rules.
\newblock \emph{Linguistic Inquiry}, 2(1):21--35.

\bibitem[{Kadmon and Landman(1993)}]{kadmon1993any}
Nirit Kadmon and Fred Landman. 1993.
\newblock \href {https://doi.org/10.1007/bf00985272} {Any}.
\newblock \emph{Linguistics and Philosophy}, 16(4):353--422.

\bibitem[{Kim and Sells(2008)}]{kim2008syntax}
Jong-Bok Kim and Peter Sells. 2008.
\newblock \emph{English Syntax: An Introduction}.
\newblock CSLI Publications.

\bibitem[{Kingma and Ba(2014)}]{kingma2014adam}
Diederik~P. Kingma and Jimmy Ba. 2014.
\newblock Adam: A method for stochastic optimization.
\newblock In \emph{Proceedings of the 3rd International Conference on Learning
  Representations}.

\bibitem[{Kiros et~al.(2015)Kiros, Zhu, Salakhutdinov, Zemel, Urtasun,
  Torralba, and Fidler}]{kiros2015skip}
Ryan Kiros, Yukun Zhu, Ruslan~R. Salakhutdinov, Richard Zemel, Raquel Urtasun,
  Antonio Torralba, and Sanja Fidler. 2015.
\newblock Skip-thought vectors.
\newblock In \emph{Advances in Neural Information Processing Systems}, pages
  3294--3302.

\bibitem[{Lau et~al.(2016)Lau, Clark, and Lappin}]{lau2016cognitive}
Jey~Han Lau, Alexander Clark, and Shalom Lappin. 2016.
\newblock Grammaticality, acceptability, and probability: {A} probabilistic
  view of linguistic knowledge.
\newblock \emph{Cognitive Science}, 41(5):1202--1241.

\bibitem[{Lawrence et~al.(2000)Lawrence, Giles, and
  Fong}]{lawrence2000grammatical}
Steve Lawrence, C.~Lee Giles, and Sandiway Fong. 2000.
\newblock Natural language grammatical inference with recurrent neural
  networks.
\newblock \emph{IEEE Transactions on Knowledge and Data Engineering},
  12(1):126--140.

\bibitem[{Levin(1993)}]{levin1993verb}
Beth Levin. 1993.
\newblock \emph{English {V}erb {C}lasses and {A}lternations: {A} preliminary
  investigation}.
\newblock University of Chicago Press.

\bibitem[{Linzen et~al.(2016)Linzen, Dupoux, and
  Goldberg}]{linzen2016assessing}
Tal Linzen, Emmanuel Dupoux, and Yoav Goldberg. 2016.
\newblock Assessing the ability of {LSTM}s to learn syntax-sensitive
  dependencies.
\newblock \emph{Transactions of the Association for Computational Linguistics},
  4:521--535.

\bibitem[{Linzen and Oseki(2018)}]{linzen2018reliability}
Tal Linzen and Yohei Oseki. 2018.
\newblock The reliability of acceptability judgments across languages.
\newblock \emph{Glossa: a journal of general linguistics}, 3(1).

\bibitem[{Mahowald et~al.(2016)Mahowald, Graff, Hartman, and
  Gibson}]{mahowald2016snap}
Kyle Mahowald, Peter Graff, Jeremy Hartman, and Edward Gibson. 2016.
\newblock {SNAP} judgments: A small n acceptability paradigm ({SNAP}) for
  linguistic acceptability judgments.
\newblock \emph{Language}, 92(3):619--635.

\bibitem[{Marvin and Linzen(2018)}]{marvin2018targeted}
Rebecca Marvin and Tal Linzen. 2018.
\newblock Targeted syntactic evaluation of language models.
\newblock In \emph{Proceedings of the 2018 Conference on Empirical Methods in
  Natural Language Processing}, pages 1192--1202.

\bibitem[{Matthews(1975)}]{matthews1975correlation}
Brian~W. Matthews. 1975.
\newblock Comparison of the predicted and observed secondary structure of t4
  phage lysozyme.
\newblock \emph{Biochimica et Biophysica Acta (BBA)-Protein Structure},
  405(2):442--451.

\bibitem[{Miller(2002)}]{miller2002syntax}
Jim Miller. 2002.
\newblock \emph{An Introduction to English Syntax}.
\newblock Edinburgh University Press.

\bibitem[{Pennington et~al.(2014)Pennington, Socher, and
  Manning}]{pennington2014glove}
Jeffrey Pennington, Richard Socher, and Christopher Manning. 2014.
\newblock {GloVe}: Global vectors for word representation.
\newblock In \emph{Proceedings of the 2014 Conference on Empirical Methods in
  Natural Language Processing}, pages 1532--1543.

\bibitem[{Peters et~al.(2018)Peters, Neumann, Iyyer, Gardner, Clark, Lee, and
  Zettlemoyer}]{peters2018elmo}
Matthew Peters, Mark Neumann, Mohit Iyyer, Matt Gardner, Christopher Clark,
  Kenton Lee, and Luke Zettlemoyer. 2018.
\newblock Deep contextualized word representations.
\newblock In \emph{Proceedings of the 2018 Conference of the North American
  Chapter of the Association for Computational Linguistics: Human Language
  Technologies, Volume 1 (Long Papers)}, volume~1, pages 2227--2237.

\bibitem[{Pullum and Scholz(2002)}]{pullum2002empirical}
Geoffrey~K. Pullum and Barbara~C. Scholz. 2002.
\newblock Empirical assessment of stimulus poverty arguments.
\newblock \emph{The Linguistic Review}, 18(1-2):9--50.

\bibitem[{Rappaport~Hovav and Levin(2008)}]{hovav2008dative}
Malka Rappaport~Hovav and Beth Levin. 2008.
\newblock The {E}nglish dative alternation: The case for verb sensitivity.
\newblock \emph{Journal of Linguistics}, 44(1):129--167.

\bibitem[{Ross(1967)}]{ross1967constraints}
John~Robert Ross. 1967.
\newblock \emph{Constraints on Variables in Syntax.}
\newblock Ph.D. thesis, MIT.

\bibitem[{Sag(1997)}]{sag1997relativeclause}
Ivan~A. Sag. 1997.
\newblock English relative clause constructions.
\newblock \emph{Journal of Linguistics}, 33(2):431--483.

\bibitem[{Sag et~al.(1985)Sag, Gazdar, Wasow, and
  Weisler}]{sag1985coordination}
Ivan~A. Sag, Gerald Gazdar, Thomas Wasow, and Steven Weisler. 1985.
\newblock Coordination and how to distinguish categories.
\newblock \emph{Natural Language \& Linguistic Theory}, 3(2):117--171.

\bibitem[{Sag et~al.(2003)Sag, Wasow, and Bender}]{sag2003syntactic}
Ivan~A. Sag, Thomas Wasow, and Emily~M. Bender. 2003.
\newblock \emph{Syntactic Theory: A Formal Introduction}, 2 edition.
\newblock CSLI Publications.

\bibitem[{Sch{\"u}tze(1996)}]{schutze1996empirical}
Carson~T. Sch{\"u}tze. 1996.
\newblock \emph{The Empirical Base of Linguistics: Grammaticality Judgments and
  Linguistic Methodology}.
\newblock University of Chicago Press.

\bibitem[{Shi et~al.(2016)Shi, Padhi, and Knight}]{shi2016syntax}
Xing Shi, Inkit Padhi, and Kevin Knight. 2016.
\newblock Does string-based neural {MT} learn source syntax?
\newblock In \emph{Proceedings of the 2016 Conference on Empirical Methods in
  Natural Language Processing}, pages 1526--1534.

\bibitem[{Sportiche et~al.(2013)Sportiche, Koopman, and
  Stabler}]{sportiche2013introduction}
Dominique Sportiche, Hilda Koopman, and Edward Stabler. 2013.
\newblock \emph{An Introduction to Syntactic Analysis and Theory}.
\newblock John Wiley \& Sons.

\bibitem[{Sprouse and Almeida(2012)}]{sprouse2012adger}
Jon Sprouse and Diogo Almeida. 2012.
\newblock Assessing the reliability of textbook data in syntax: Adger's {C}ore
  {S}yntax.
\newblock \emph{Journal of Linguistics}, 48(3):609--652.

\bibitem[{Sprouse and Almeida(2017)}]{sprouse2017setting}
Jon Sprouse and Diogo Almeida. 2017.
\newblock Setting the empirical record straight: Acceptability judgments appear
  to be reliable, robust, and replicable.
\newblock \emph{Behavioral and Brain Sciences}, 40.

\bibitem[{Sprouse et~al.(2013)Sprouse, Sch{\"u}tze, and
  Almeida}]{sprouse2013li}
Jon Sprouse, Carson~T. Sch{\"u}tze, and Diogo Almeida. 2013.
\newblock A comparison of informal and formal acceptability judgments using a
  random sample from {L}inguistic {I}nquiry 2001--2010.
\newblock \emph{Lingua}, 134:219--248.

\bibitem[{Wagner et~al.(2009)Wagner, Foster, and van
  Genabith}]{wagner2009grammaticality}
Joachim Wagner, Jennifer Foster, and Josef van Genabith. 2009.
\newblock Judging grammaticality: Experiments in sentence classification.
\newblock \emph{CALICO Journal}, 26(3):474--490.

\bibitem[{Wang et~al.(2018)Wang, Singh, Michael, Hill, Levy, and
  Bowman}]{wang2018glue}
Alex Wang, Amanpreet Singh, Julian Michael, Felix Hill, Omer Levy, and Samuel
  Bowman. 2018.
\newblock {GLUE}: {A} multi-task benchmark and analysis platform for natural
  language understanding.
\newblock In \emph{Proceedings of the 2018 EMNLP Workshop BlackboxNLP:
  Analyzing and Interpreting Neural Networks for NLP}, pages 353--355.

\bibitem[{Wilcox et~al.(2018)Wilcox, Levy, Morita, and Futrell}]{wilcox2018rnn}
Ethan Wilcox, Roger Levy, Takashi Morita, and Richard Futrell. 2018.
\newblock What do {RNN} language models learn about filler--gap dependencies?
\newblock In \emph{Proceedings of the 2018 EMNLP Workshop BlackboxNLP:
  Analyzing and Interpreting Neural Networks for NLP}, pages 211--221.

\bibitem[{Wilcox et~al.(2019)Wilcox, Qian, Futrell, Ballesteros, and
  Levy}]{wilcox2019structural}
Ethan Wilcox, Peng Qian, Richard Futrell, Miguel Ballesteros, and Roger Levy.
  2019.
\newblock Structural supervision improves learning of non-local grammatical
  dependencies.
\newblock In \emph{Proceedings of the 2019 Conference of the North American
  Chapter of the Association for Computational Linguistics: Human Language
  Technologies}, pages 3302--3312.

\bibitem[{Williams(1980)}]{williams1980predication}
Edwin Williams. 1980.
\newblock Predication.
\newblock \emph{Linguistic Inquiry}, 11(1):203--238.

\end{thebibliography}
\bibliographystyle{acl_natbib}

\end{document}